\documentclass[sigconf]{acmart}
\AtBeginDocument{%
  }

\usepackage{booktabs}
\usepackage{tabularx}
\usepackage{graphicx}
\usepackage{xcolor}
\usepackage{multirow}
\usepackage{color}
\usepackage{colortbl}
\usepackage{caption}
\usepackage{enumitem}
\usepackage{subcaption}
\usepackage{balance}
\usepackage[english]{babel}
\usepackage[utf8x]{inputenc}
\usepackage[colorinlistoftodos]{todonotes}

\usepackage{subfiles}

\definecolor{myred}{RGB}{255, 99, 71}
\definecolor{mygreen}{RGB}{0, 255, 0}
\definecolor{myblue}{RGB}{100, 216, 255}

\newcommand{\paratitle}[1]{\vspace{1ex}\noindent \textbf{#1}}

\copyrightyear{2025}
\acmYear{2025}
\setcopyright{cc}
\setcctype{by}
\acmConference[KDD '25]{Proceedings of the 31st ACM SIGKDD Conference on Knowledge Discovery and Data Mining V.2}{August 3--7, 2025}{Toronto, ON, Canada}
\acmBooktitle{Proceedings of the 31st ACM SIGKDD Conference on Knowledge Discovery and Data Mining V.2 (KDD '25), August 3--7, 2025, Toronto, ON, Canada}
\acmDOI{10.1145/3711896.3736937}
\acmISBN{979-8-4007-1454-2/2025/08}

\begin{document}

\title{Enhancing Large Language Models for Mobility Analytics with Semantic Location Tokenization}


\author{Yile Chen}
\affiliation{
  \institution{Nanyang Technological University}
  \country{Singapore}
}
\email{yile001@e.ntu.edu.sg}

\author{Yicheng Tao}
\affiliation{
  \institution{Southern University of Science and Technology}
  \city{Shenzhen}
  \country{China}
}
\email{12112003@mail.sustech.edu.cn}
\authornote{Work done during an internship at Nanyang Technological University}

\author{Yue Jiang}
\affiliation{
  \institution{Nanyang Technological University}
  \country{Singapore}
}
\email{yue013@e.ntu.edu.sg}

\author{Shuai Liu}
\affiliation{
  \institution{Nanyang Technological University}
  \country{Singapore}
}
\email{shuai004@e.ntu.edu.sg}

\author{Han Yu}
\affiliation{
  \institution{Nanyang Technological University}
  \country{Singapore}
}
\email{han.yu@ntu.edu.sg}

\author{Gao Cong}
\affiliation{
  \institution{Nanyang Technological University}
  \country{Singapore}
}
\email{gaocong@ntu.edu.sg}

\renewcommand{\shortauthors}{Yile Chen et al.}

\begin{abstract}
The widespread adoption of location-based services has led to the generation of vast amounts of mobility data, providing significant opportunities to model user movement dynamics within urban environments. Recent advancements have focused on adapting Large Language Models (LLMs) for mobility analytics. However, existing methods face two primary limitations: inadequate semantic representation of locations (i.e., discrete IDs) and insufficient modeling of mobility signals within LLMs (i.e., single templated instruction fine-tuning).
To address these issues, we propose QT-Mob, a novel framework that significantly enhances LLMs for mobility analytics. QT-Mob introduces a location tokenization module that learns compact, semantically rich tokens to represent locations, preserving contextual information while ensuring compatibility with LLMs. Furthermore, QT-Mob incorporates a series of complementary fine-tuning objectives that align the learned tokens with the internal representations in LLMs, improving the model's comprehension of sequential movement patterns and location semantics. The proposed QT-Mob framework not only enhances LLMs' ability to interpret mobility data but also provides a more generalizable approach for various mobility analytics tasks. Experiments on three real-world dataset demonstrate the superior performance in both next-location prediction and mobility recovery tasks, outperforming  existing deep learning and LLM-based methods.

\end{abstract}




\maketitle

\vspace{-2mm}
\section{Introduction}
With the prevalence of geopositioning technologies and location-based services, vast amounts of location footprints have been generated through both passive sources (e.g., cellular signals) or active interactions (e.g., check-ins). These massive mobility records serve as valuable data to understand human movement dynamics within urban environments, thus offering significant benefits across various downstream scenarios~\cite{mobility_survey, ssl_geospatial_survey}. For example, mobility analytics can enhance  location recommendation services~\cite{POI_survey1} for businesses, while also improving the quality of urban monitoring~\cite{stream_miao} to support more informed urban management strategies.

Modeling mobility data has been widely studied over the past decade, particularly in the topics of next location prediction and mobility recovery. The core objective is to capture sequential patterns while effectively considering complex spatio-temporal dependencies inherent in mobility records. To achieve this, early studies focused on statistical methods, such as matrix factorization and Markov chain~\cite{Markov3_KDD15, Markov4_Ubicomp12}, to model user movement behavior. More recently, deep learning-based models, including sequential models such as Recurrent Neural Networks (RNN)~\cite{Deepmove_WWW18, mobility_AAAI20, mobility_TKDE22} and Transformer architecture~\cite{mobility_trans_WWW21, mobility_trans_SIGIR22, mobility_trans_KDD24}, and models based on Graph Neural Networks (GNNs)~\cite{mobility_gnn_CIKM20,mobility_gnn_KDD22,mobility_gnn_SIGIR23}, have been adapted to achieve significant improvements over statistical methods. These models typically integrate mobility semantics, such as spatio-temporal context and location attributes, by encoding them as vector representations via embedding tables. These representations are then incorporated into the design of sequential learning frameworks to enhance predicative performance. 
 
\begin{figure}[tbp]
\centering
\includegraphics[width=0.98\linewidth]{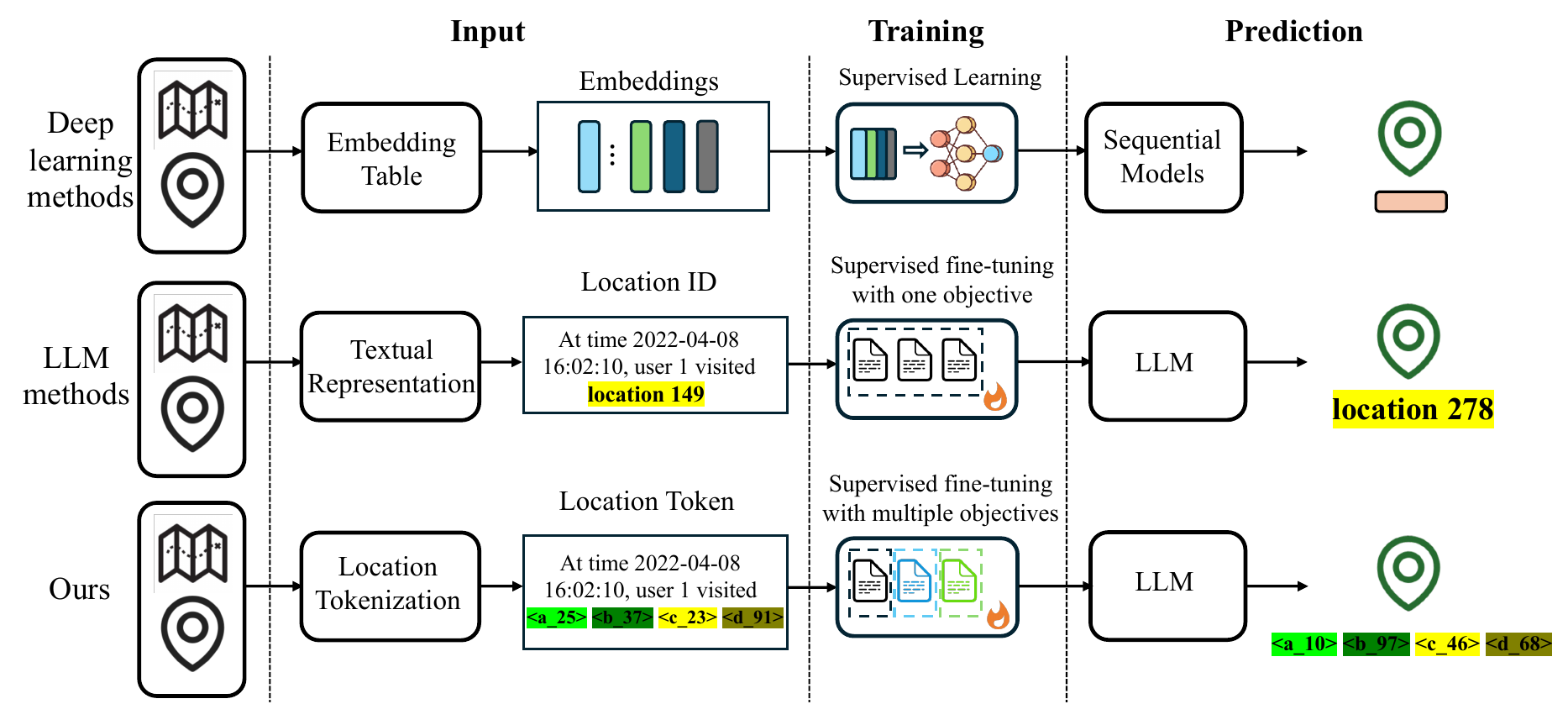}
\vspace{-2mm}
\caption{Comparison between QT-Mob and previous methods for mobility analytics.}
\label{fig:intro}
\vspace{-6mm}
\end{figure}

The rich semantic information in mobility records, as modeled in deep learning-based models, such as location coordinates, categories, and timestamps, is inherently rooted in textual data. Such textual information is commonly stored in language corpora, which Large Language Models (LLMs) have been extensively trained on to develop reasoning capabilities~\cite{LLM_survey}. Consequently, LLMs naturally possess a comprehensive understanding of general geospatial knowledge~\cite{GeoLLM, UrbanFoundationModel}. This has led to increasing interest in adapting LLMs for mobility data for downstream analytical tasks. For example, recent studies have explored designing prompts~\cite{AgentMove, LLMprompt2} and agent frameworks~\cite{LLMUrbanResidence} to interact with LLMs for predicting or simulating mobility trajectories. Besides, LLMs are also employed to replace previous sequential models~\cite{MobilityLLM}, or fine-tuned using instruction-tuning paradigms~\cite{LLMPOI1, GenUP}.

Despite the demonstrated effectiveness of LLM-based methods in tackling mobility records from a textual perspective and the promising results achieved, existing methods adapting LLMs for mobility analytics face two critical limitations. \underline{First}, they follow the practices of earlier deep learning-based models by representing locations as discrete IDs (e.g., ``location 124”) in location sequence descriptions provided in prompts~\cite{LLMPOI1, LLMprompt2}. While categories and timestamps are usually supplemented to enrich the descriptions, the use of discrete IDs fails to encode the semantic richness and contextual information of locations. 
\underline{Second}, LLMs are not  adapted to fully comprehend the domain-specific knowledge embedded in mobility data. Interacting with fixed LLMs through prompts as in~\cite{AgentMove, LLMprompt2} introduces no additional knowledge beyond the general geospatial concepts already encoded in LLMs. Moreover, fine-tuning LLMs on a single templated instruction dataset (e.g., solely for next location prediction) captures mobility knowledge from only a single perspective. This limited scope usually results in a biased model specialized for a single task, and insufficient to develop a comprehensive understanding of the insights within mobility data.

To address these limitations, we propose a novel framework named \textbf{QT-Mob}, to fully trigger the capabilities of LLMs for mobility analytics, achieving superior performance compared to previous deep learning-based and LLM-based methods. The comparison between QT-Mob and previous methods is illustrated in Figure~\ref{fig:intro}. QT-Mob consists of two key components.  
First, it introduces a location tokenization module, which learns location tokens that encode rich location semantic and contextual information via hierarchical vector quantization techniques. This design allows each location to be represented as a fixed and compact sequence of a few discrete tokens (e.g., 4 tokens), while preserving the intrinsic and contextual properties, which are initially described by lengthy textual descriptions. Furthermore, locations with higher semantic similarity are assigned more similar token sequences, ensuring semantic coherence in the learned tokens. By introducing only a small number of newly introduced tokens, this module maintains the discrete nature of LLM inputs, facilitating seamless integration with LLMs and minimizing the burden for subsequent fine-tuning.
Second, building on standard instruction-tuning paradigms, QT-Mob incorporates a set of complementary objectives to fine-tune LLMs. These objectives are designed to not only enhance the model's ability to comprehend sequential patterns and location semantics, but also refine the alignment between the derived location tokens and the LLM's internal representations. As a result, the fine-tuned LLMs gain a deeper comprehension of mobility data, capturing multifaceted mobility insights from diverse aspects. This enhanced understanding enables the model to be effectively applied to general mobility analytical tasks, including next location prediction and mobility recovery.    

Our main contributions can be summarized as follows:
\vspace{-1mm}
\begin{itemize}[leftmargin=*]
    \item We propose QT-Mob, a novel framework that adapts LLMs to understand the semantic richness and contextual information of mobility records within textual space for the first time, achieving more effective performance in mobility analytics. 
    \item We propose to encode semantics of locations by discrete tokens while maintaining feasible compatibility for LLMs. Besides, we fine-tune LLMs through a set of carefully designed objectives that infuse mobility-specific knowledge and ensure the coherent integration of location tokens into LLMs.    
    \item We conduct extensive experiments on three real-world mobility datasets on both next location prediction and mobility recovery tasks. Experimental results demonstrate that our proposed framework outperforms state-of-the-art methods.

\end{itemize}

\vspace{-2mm}
\section{Related Work}

\subsection{Next Location Prediction}
Next location prediction serves as a fundamental problem in modeling user preferences for location-based services. Traditional methods are primarily based Markov chain models to capture transition probabilities between locations~\cite{Markov3_KDD15, Markov4_Ubicomp12}, or metric embedding techniques to model location similarities within location sequences~\cite{mobility_IJCAI15, mobility_KDD16}. With the advent of deep learning, sequential models have been widely applied to this task. These models typically incorporate  spatial and temporal contexts (e.g., geographical distances and time intervals) into the framework, adapting models such as RNN~\cite{Deepmove_WWW18, mobility_AAAI20, mobility_TKDE22} and Transformer~\cite{mobility_trans_WWW21, mobility_trans_SIGIR22, mobility_trans_KDD24}. Besides, to capture high-order user-location collaborative signals, several studies propose to aggregate the mobility records into the graph structure, and employ GNN models to learn transition patterns~\cite{mobility_gnn_CIKM20,mobility_gnn_KDD22,mobility_gnn_SIGIR23, mobility_GNN_SIGIR24}. More recently, advancements of LLMs have inspired their application to mobility analytics. For example, LLM-Move~\cite{LLMprompt2} and AgentMove~\cite{AgentMove} organize  prompts to define  next location prediction task and  objectives, converting  mobility records into textual format for fixed LLMs (e.g., ChatGPT) to generate predictions. LLMob~\cite{LLMUrbanResidence} adopts an agent framework to simulate user activity patterns and motivations, generating mobility records for various user profiles. For fine-tuning paradigm, MobilityLLM~\cite{MobilityLLM} replaces the Transformer architecture with LLMs and fine-tunes the model to several mobility analytical tasks. LLM4POI~\cite{LLMPOI1} fine-tunes LLMs with a instruction tuning dataset constructed from  mobility records enriched with contextual information, such as categories and similar mobility sequences. Building on this, GenUP~\cite{GenUP} incorporates user profiles into system prompts to enhance the performance with user-specific personality traits.  In contrast, our proposed QT-Mob effectively addresses the limitations of previous methods, namely the reliance on numerical ID representations and the superficial adaptation of LLM to mobility comprehension, achieving state-of-the-art performances for mobility analytics.    

\vspace{-3mm}
\subsection{Mobility Recovery}
The essence of mobility recovery is to reconstruct high-quality trajectories from sparse trajectories with unknown locations. When road network information is available, this problem is formulated as a task involving additional structural constraints or conditions~\cite{KDD21_recovery, KDD22_camerarecovery}. In the absence of such external information, the problem is referred to as network-free mobility recovery, which aligns with the focus of this study. Early studies propose to predict unknown locations through calibration with pre-defined anchor points~\cite{SIGMOD13_calibration}, or Markov models~\cite{ICDE11_recovery, CIKM16_recovery}. However, these approaches simply consider low-order transition patterns in a trajectory.
To overcome the limitations, deep learning methods have been increasingly introduced. For example, Bi-STDDP~\cite{AAAI19_recovery} employs RNN to model the local location correlations with a fixed window size using metric embedding techniques. DHTR~\cite{TKDE21_recovery} adopts a sequence-to-sequence framework equipped with spatio-temporal attentions and Kalman filter to capture transition patterns. AttnMove~\cite{AttnMove} and PeriodicMove~\cite{CIKM_recovery}  employ self-attention mechanisms both within and across trajectories to recover unobserved locations. TrajBERT~\cite{TrajBERT} applies BERT-style~\cite{BERT} pre-training strategies to encode the incomplete trajectory and incorporates spatial and temporal refinement modules to enhance accuracy. In this context, our proposed QT-Mob provides an effective way  of capturing mobility patterns from diverse objectives, offering adaptability to mobility recovery task.

\vspace{-3mm}
\subsection{LLMs for Geospatial Analytics}
The adaptation of LLMs has emerged as a significant trend across
various domains, yielding promising results in diverse data modalities, including time series, images, and graphs~\cite{LLMGraph, timeLLM}. Building on this success, LLMs have also been adopted in geospatial analytics due to their understanding of general geospatial knowledge~\cite{GeoAI2}. For example, 
GeoLLM~\cite{GeoLLM} harnesses LLMs to encode regional features from textual prompts, such as addresses and nearby districts, and utilizes this information for LLMs to generate targeted response variables for several socioeconomic indicators. Moreover, fine-tuning LLMs with fine-grained urban knowledge, such as flight schedules, points of interest (POIs), and hotel data, has enabled conversational applications in tasks like route recommendation~\cite{ICML24_GeoAI}, location search~\cite{LAMP}, and urban planning~\cite{PlanGPT}. In traffic management, LLMLight~\cite{LLMLight} employs LLMs as agents to manege traffic signal control with interpretability, and generalization ability.  UrbanGPT~\cite{UrbanGPT} constructs prompts that combine
textual descriptions with representations from spatiotemporal learning methods, enabling LLMs to perform effective traffic forecasting. In this study, we build on the demonstrated capabilities of LLMs in geospatial analytics, specifically addressing the limitations of prior methods in mobility analytics. 

\section{Problem Formulation}
In this section, we present key definitions used throughout this paper, and formulate the research problem. Let $U=\{u_1, u_2, ..., u_M\}$  be a set of users, and $L=\{l_1, l_2,...,l_N\}$ be a set of locations. Each location $l_i \in L$ is associated with a name $m$, a geographical coordinates $(x,y)$ and a category $c$.

\begin{definition}
 (\textbf{Mobility Record}). Mobility records represent the locations visited by users in a geospatial space. Each mobility record is defined as a tuple $r = (u, t, l)$, denoting a user $u$ visits location $l$ at time $t$.
\end{definition}

\begin{definition}
(\textbf{Mobility Trajectory}). A mobility trajectory is defined as an ordered sequence of mobility records generated by a user $u$, denoted as $T_{u} = (r_{1}^{u}, r_{2}^{u},..., r_{n}^{u})$, where $t_{i+1}- t_{i} < \Delta t$. Each $r_{i}^{u}$ corresponds to a mobility record, and consecutive records in the trajectory occur within a specified time interval (e.g., $\Delta t=24$ hours). 
\end{definition}

Our target is to adapt LLMs to effective understand and reason about mobility trajectories. The expectation is that the adapted LLMs essentially capture the complex mobility insights, such as spatio-temporal dependencies and sequential patterns, enabling their application to analytical tasks that require such nuanced comprehension. To demonstrate this capability, we evaluate the performance on two representative tasks: \textit{next location prediction} and \textit{mobility recovery}.

Given the historical mobility trajectories $\mathcal{T}_u = \{T_{u}^{1}, T_{u}^{2}, ..., T_{u}^{k}\}$, the two tasks can be formulated as:

\begin{definition}
(\textbf{Next Location Prediction}). For the current mobility trajectory $T_{u}^{k+1} = (r_{1}^{u}, r_{2}^{u},...r_{j}^{u})$, the objective is to predict the most likely next location $l_{j+1}$ visited by the user $u$ at the subsequent timestamp $t_{j+1}$. 
\end{definition}

\begin{definition}
(\textbf{Mobility Recovery}). For the current trajectory $T_{u}^{k+1} = (r_{1}^{u}, r_{2}^{u},...r_{j}^{u})$, where some mobility records have missing location information (i.e., $\exists r_{i}^{u}, r_{i}^{u}=(u, t_{i}, \varnothing)$), the objective is to recover the missing locations $l_{i}\in r_{i}^{u}$ for these records in $T_{u}^{k+1}$. 
\end{definition}

\vspace{-2mm}
\section{Methodology}
\begin{figure*}
    \centering
    \includegraphics[width=1\linewidth]{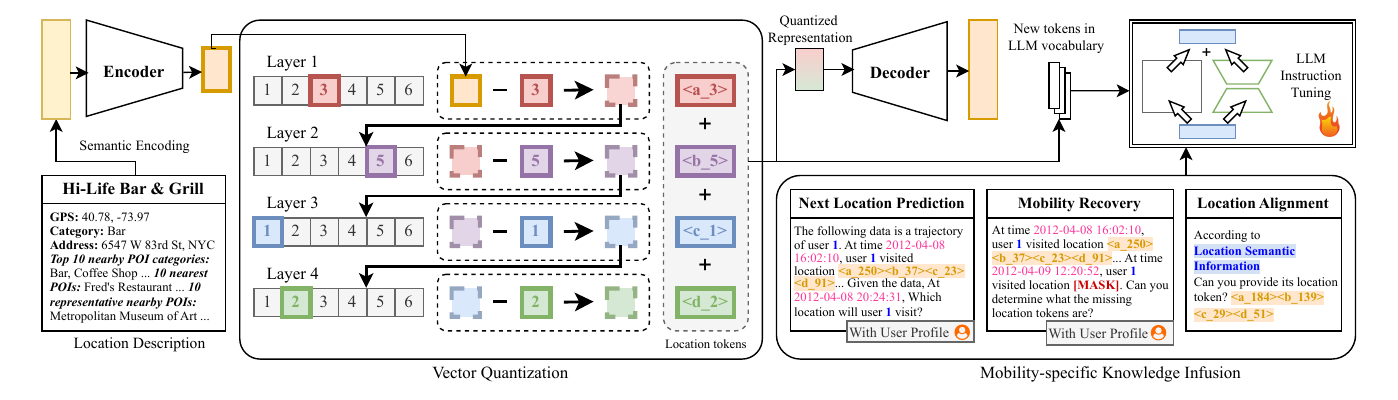}
    \vspace{-7mm}
    \caption{Overview of proposed QT-Mob framework}
    \label{fig:qtmob}
    \vspace{-4mm}
\end{figure*}

In this section, we introduce our QT-Mob framework, as illustrated in Figure~\ref{fig:qtmob}. We first present an overview of QT-Mob, followed by a detailed explanation of the location tokenization module and the infusion of mobility-specific knowledge into LLMs. Our code is available at: \href{https://github.com/shadowfall09/QT-Mob}{\texttt{https://github.com/shadowfall09/QT-Mob}}.

\vspace{-2mm}
\subsection{Framework Overview}

The core idea of QT-Mob is to encode the semantic richness of locations into compact and discrete tokens while ensuring their seamless integration into LLMs, thereby enhancing the model’s ability to comprehend mobility trajectories. To achieve this, QT-Mob consists of a two-stage process. 

First, we encode textual descriptions of each location, including attributes such as title, category, coordinates, and nearby locations, into a set of specialized location tokens via hierarchical vector quantization techniques. The generated tokens preserve the intrinsic and contextual properties of locations that vanilla location IDs fail to convey, maintaining hierarchical similarities where similar location share common token prefixes. This transformation effectively condenses lengthy textual descriptions into a short tokenized sequence, and produces limited additional tokens to the vocabulary of LLMs. This significantly reduces the difficulty for the alignment of LLMs with mobility data in the subsequent fine-tuning stage.

Second, rather than relying on a single templated instruction-tuning objective for next location prediction, we introduce multiple complementary training objectives to enable LLMs to develop a deeper understanding of mobility data. Apart from the next location prediction objective, we incorporate token-to-text and text-to-token location alignment objectives, which ensure that the specialized location tokens are effectively integrated into the textual space of LLMs. Besides, we introduce a mobility recovery objective designed to consider bidirectional dependencies, which are not emphasized in next location prediction task. These objectives not only bridge the LLMs with the derived location tokens, but also allow the model to capture the multifaceted insights in mobility trajectories. Through this process, QT-Mob enhances LLMs by injecting the awareness of location semantics and the mobility-specific knowledge, thereby improving their effectiveness in mobility analytical tasks.

\vspace{-2mm}
\subsection{Location Tokenization Module}
To enhance the capability of LLMs in representing locations, we avoid using vanilla location IDs~\cite{GenUP, LLMPOI1}, as they lack inherent  semantic information and fail to capture the contextual relationships between locations. Instead, we adopt a tokenization strategy based on RQ-VAE~\cite{RQ-VAE, ICDE24_recsys}, which encodes location descriptions into a sequence of discrete tokens. These tokens are optimized in a coarse-to-fine manner, which embeds hierarchical semantic meaning. The tokenization process consists of two key steps: semantic encoding and vector quantization.

\paratitle{Semantic Encoding}.  To comprehensively capture the semantic meaning of each location, we construct a location description $D$ that contains both intrinsic attributes and the contextual information, providing a richer summarization compared to simple location IDs. In particular, the intrinsic attributes include location name, categories, geographical coordinates, geohash code and address. In addition, the contextual information includes the top-K nearest POIs to capture spatial relationships, the most frequently visited POIs within a certain distance (e.g., 2 km) to exhibit representative landmarks, and the top-K visited categories to reflect the aggregated functional patterns in the surrounding region. An example of a structured location description can be found in Appendix~\ref{app:loc-description}.
 Given this textual description, we derive the initial location representation $\mathbf{s}$ using a pre-trained language models (e.g., Llama), as a semantic encoder: 

 \begin{equation}\label{eq:encoder}
      \mathbf{s}=  \operatorname{Encoder}(D) \in \mathbb{R}^{d}.
 \end{equation}

\begin{figure*}[htbp]
    \centering
    \includegraphics[width=1\linewidth]{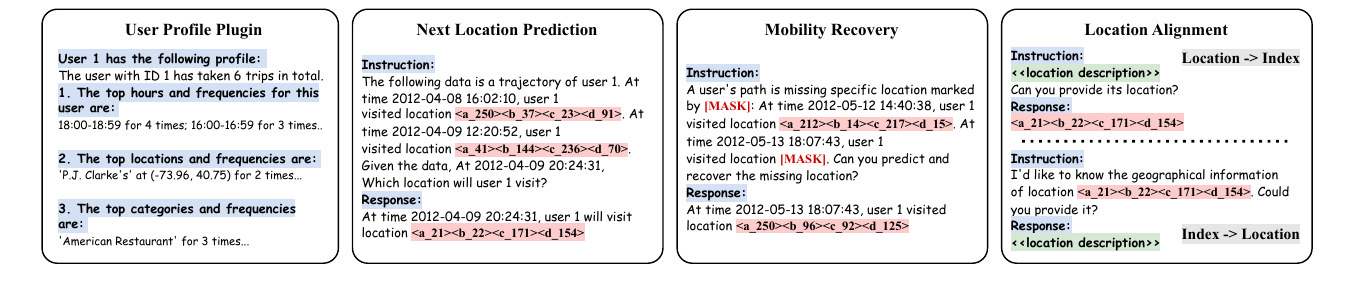}
    \vspace{-6mm}
    \caption{Instruction tuning datasets constructed with multiple objectives}
    \label{fig:prompt}
\vspace{-3mm}
\end{figure*}

\paratitle{Vector Quantization}. Given the location representation $\mathbf{s}$ , we aim to generate a sequence of discrete tokens that align  with the input format of LLMs while preserving the semantic integrity of locations. To this end, we employ RQ-VAE~\cite{RQ-VAE, ICDE24_recsys}, which recursively quantizes the representation through multiple layers of codebooks, thus resulting in a hierarchical encoding of location semantics. Specifically, for each quantization level $l \in [1, L]$, we define a codebook $\mathcal{C}^{l}=\{\mathbf{v}_{i}^{l}\}_{i=1}^{K}$, where each $\mathbf{v}_{i}^{l}$ corresponds to a learnable token representation, and $K$ denotes the codebook size. Then the quantization process at each level can be formulated as:
\vspace{-2mm}
\begin{equation}
\begin{aligned}
c_l &= \underset{i}{\arg \min }\left\|\boldsymbol{r}_{l-1}-\mathbf{v}_{i}^{l}\right\|^2  \\
\boldsymbol{r}_l &= \boldsymbol{r}_{l-1}-\mathbf{v}_{c_l}^{l}
\end{aligned}
\end{equation}
where $c_i$ is the assigned code index (i.e., token) at the $l-$layer of the codebook, $\boldsymbol{r}_{l-1}$ is the residual embedding from the previous layer, initialized as $\boldsymbol{r}_0 = \mathbf{s}$. 
This process enables hierarchical encoding, as the original location representation $\mathbf{s}$ is progressively decomposed into a sequence of $L$ tokens, each corresponding to the closest match in the respective codebook layer.
After this process, we obtain the quantized indices $\mathbf{c}=[c_1, ..., c_L]$, and the corresponding quantized representation $\hat{\mathbf{z}}=\sum_{l=1}^L \boldsymbol{v}_{c_l}^l$. For training, we employ a reconstruction loss, where $\hat{\mathbf{z}}$ is utilized to reconstruct the initial location representation $\mathbf{s}$ through MLP decoder:

\vspace{-2mm}
\begin{equation}
\mathcal{L}_{\mathrm{Rec}} =\|\mathbf{s}-\operatorname{MLP}( \hat{\mathbf{z}})\|_2^2 
\end{equation}

Since the index selection operation via $argmin$ is not differentiable, we assign the gradients following $argmin$ to preceding steps. This enables end-to-end training of both the encoding and the decoding processes~\cite{discrete, discrete2}. Furthermore, to optimize the token representations, we enforce them to be close to the encoded residual embeddings, and integrate a residual quantization loss as follows:

\begin{equation}
    \mathcal{L}_{\mathrm{RQ}} =\sum_{l=1}^L\left\|\operatorname{sg}\left[\boldsymbol{r}_{l}\right]-\boldsymbol{v}_{c_l}^l\right\|_2^2+\alpha\left\|\boldsymbol{r}_{l}-\operatorname{sg}\left[\boldsymbol{v}_{c_l}^l\right]\right\|_2^2
\end{equation}
where $\operatorname{sg}[\cdot]$ denotes the stop-gradient operation, and $\alpha$ is a weight to balance the optimization of the token representations and the encoder~\cite{discrete}. The final objective function combines the reconstruction loss and the quantization loss to jointly optimize the encoding, the decoding, and the codebook components:

\vspace{-2mm}
\begin{equation}
    \mathcal{L} =\mathcal{L}_{\mathrm{Rec}}+\mathcal{L}_{\mathrm{RQ}}
\end{equation}

\subsection{Infusing Mobility-Specific Knowledge}\label{subsec:sft}

Building on the semantic location tokens derived in the previous module, we aim to infuse mobility-specific knowledge into LLMs with these tokens, enhancing the LLMs' understanding of mobility data beyond the inherently pretrained geospatial knowledge. To achieve this, we adopt an instruction-tuning paradigm, which fine-tunes LLM with high-quality, domain-specific datasets. Since the location tokens are essentially out-of-vocabulary tokens that LLMs do not recognize, we deviate from single templated instruction-tuning dataset used in prior studies~\cite{LLMPOI1, GenUP}, which focuses solely on next location prediction. Instead, we formulate a set of complementary fine-tuning objectives that align the location tokens with the internal representations of LLMs in textual space. This enables LLMs to develop a deeper understanding of mobility insights, allowing for more effective adaptation to downstream applications. Figure~\ref{fig:prompt} provides an overview of these objectives, with instances illustrating their implementation.

\paratitle{Next Location Prediction}. 
As sequential patterns are fundamental for mobility modeling, we adopt next location prediction as the primary fine-tuning objective, consistent with prior research on next location prediction. Specifically, we construct instruction-based prompts using the user’s current mobility trajectory and fine-tune LLMs to predict the most likely next location. Each mobility record is represented as timestamps in natural language format, and the semantic location tokens in chronological order. As user's historical mobility trajectories provide valuable signals for prediction, we integrate them into the prompts for enhanced prediction accuracy. However, directly incorporating full historical records often results in excessively long input sequences, which may hinder model effectiveness and training efficiency. To address this, we propose to generate user profiles based on statistical features extracted from historical mobility trajectories.  These features include the most frequently visited hours, locations, and categories, along with their corresponding frequencies. The resulting user profile is appended before the current trajectory, serving as a concise yet informative exhibition of general user preferences. 

\paratitle{Mobility Recovery}.
The next location prediction objective mainly focuses on modeling causal dependencies for predicting the last location in a sequence. While effective for capturing sequential patterns, this objective does not fully account for the bidirectional dependencies inherent in trajectories. To further enhance the comprehension of mobility insights, we introduce a mobility recovery objective. Specifically, we randomly replace a certain ratio of locations with \texttt{[MASK]} and fine-tune the LLMs to recover these unknown locations using location tokens in its response.  This allows the model to further leverage bidirectional dependencies in mobility trajectories, mitigating the emphasis of merely causal reasoning for next location prediction. By incorporating this objective, LLMs gain a more holistic understanding of mobility data, leading to enhanced capabilities and generalization for mobility analytics tasks.

\paratitle{Location Alignment}.
Although the location tokens encode hierarchical semantic meaning based on location descriptions, they are newly introduced tokens and initially exhibit weak correlations with the existing LLM textual space. To effectively integrate these location tokens into LLMs and establish coherent semantic alignments, we treat this process as a cross-modal alignment task and introduce two explicit location alignment objectives for fine-tuning. On the one hand, LLMs should be able to identify location tokens based on their associated textual descriptions. On the other hand, LLMs should also be capable of generating location descriptions from their corresponding tokens. Considering these two aspects, we construct instances where LLMs are prompted to generate location tokens given a detailed location description. In addition, we introduce instances requiring LLMs to recover full location descriptions based on input location tokens. By incorporating both objectives, LLMs are trained to seamlessly integrate location tokens into their textual representation space, enabling more effective reasoning over mobility data.

By organizing and curating instruction-tuning datasets based on these objectives, we apply supervised fine-tuning to LLMs, enabling the model to fully unlock the ability of acquiring mobility-specific knowledge for mobility analytical tasks.
\vspace{-2mm}
\section{Experiments}
\subsection{Experiment Settings}
\subsubsection{Datasets}

We conduct the experiments using three real-world datasets, namely New York City (NYC)~\cite{Foursquare},  Singapore (SG)~\cite{SG_dataset}, and Cellular trajectories (CE)~\cite{TERI}. The NYC and SG datasets are derived from Foursquare active check-ins on POIs within the respective cities from Apr 2012 to Feb 2013. The CE dataset, provided by a telecom company in Singapore, reflects mobility trajectories based on passive cellphone connections to cell towers during Nov 2022. For location representation, POIs are treated as discrete locations in the NYC and SG datasets, while in the CE dataset, geographic space is partitioned into 400-meter grid cells, each considered a location.  We preprocess the dataset by sorting the mobility records in chronological order and removing locations with fewer than 5 visits. Mobility trajectories are constructed by grouping consecutive records occurring within a 24-hour window, with trajectories containing fewer than three records being discarded. We take the first 70\% mobility trajectories of each user as training set, the subsequent 10\% as validation set, and the last 20\% as test set. The statistics of the preprocessed datasets are presented in Table \ref{tab:dataset}.

\begin{table}[t]
\centering
\caption{Dataset Statistics}
\vspace{-3mm}
\small
\begin{tabularx}{0.95\linewidth}{lXXXX}
\toprule
\textbf{Dataset} & \#Users & \#Locations & \#Records & \#Trajectories \\
\midrule
\textbf{NYC} & 1,081 & 5,765 & 66,504 & 11,297 \\
\textbf{SG} & 1,955  & 5,452 & 107,351 & 18,235 \\
\textbf{CE} & 9,700  & 3,129 & 378,514 & 46,196 \\
\bottomrule
\end{tabularx}
\label{tab:dataset}
\vspace{-6mm}
\end{table}

\vspace{-2mm}
\subsubsection{Baseline Methods}
Unlike previous studies that focus on a single task with models specifically designed for particular input formats (e.g., location IDs), QT-Mob is designed to develop a general semantic understanding of mobility data within the textual space. By leveraging this broader comprehension and enhanced flexibility, QT-Mob can be applied in several mobility analytics tasks. Specifically, we test two representative tasks: \textbf{next location prediction} and \textbf{mobility recovery}, against task-specific models.

The next location prediction task is compared with the following baseline methods:

\begin{itemize}[leftmargin=*]
    \item \textbf{FPMC} \cite{FPMC}: it is a widely used statistical method that integrates the Bayesian Personalized Ranking framework with Markov chain and matrix factorization for next location prediction.
    \item \textbf{DeepMove} \cite{Deepmove_WWW18}: it is a RNN-based model that captures sequential transitions and utilizes historical attention mechanism to learn multi-level periodicity in mobility records.
    \item \textbf{CTLE} \cite{CTLE}: it is a pre-training model that generates context-specific location representations based on the surrounding locations using a Transformer encoder.
    \item \textbf{TrajFormer} \cite{TrajFormer}: it is a transformer-based model that adopts a squeezed self-attention module, and models trajectory irregularity by introducing a continuous point embedding module.
    \item \textbf{GETNext} \cite{mobility_trans_SIGIR22}: it is a Transformer-based framework that constructs a global trajectory flow map and integrates spatiotemporal context along with time-aware category embeddings for next-location prediction.
    \item \textbf{MCLP} \cite{mobility_trans_KDD24}: it is a multi-context-aware model that includes a topic model for user preference extraction, an arrival time estimator for temporal embeddings, and a Transformer model to capture sequential mobility patterns.
    \item \textbf{PLSPL} \cite{mobility_TKDE22}: it captures long-term user preferences using attention mechanism on contextual features of locations in historical records, while modeling short-term preferences with two parallel LSTMs for location and category sequences.
     \item \textbf{STHGCN} \cite{mobility_gnn_SIGIR23}: it introduces a hypergraph Transformer to effectively combine hypergraph structures derived from inter-user and intra-user relations with spatio-temporal contexts.
    \item \textbf{LLM-Move} \cite{LLMprompt2}: it is a prompt-based framework that enables LLMs to perform next location prediction by incorporating long-term and recent user check-ins, geographic distance, and sequential transitions into structured prompts.
    \item \textbf{AgentMove} \cite{AgentMove}: it is LLM-based agentic framework for next location prediction, consisting of a spatial-temporal memory module, urban structure modeling via a world knowledge generator, and collective pattern extraction for generating predictions.
    \item \textbf{GenUP} \cite{GenUP}: it is a fine-tuned LLM-based model that extends ~\cite{LLMPOI1} by incorporating instruction-tuned datasets enriched with user profiles, including personality assessments and behavioral theories, along with historical mobility records. 
    \item \textbf{MobilityLLM} \cite{MobilityLLM}:  it is a state-of-the-art model that replaces traditional sequential models with LLMs for processing location sequences, incorporating a visiting intention memory network and human travel preference prompts to extract additional semantic information.
\end{itemize}

Apart from the primarily focused next location prediction task, we also evaluate QT-Mob on mobility recovery task to demonstrate its enhanced adaptability. We set the ratio of unknown locations to be from 20\%-50\%, and compare QT-Mob with $\textbf{LSTM}$~\cite{LSTM}, $\textbf{AttnMove}$~\cite{AttnMove}, $\textbf{TrajBERT}$~\cite{TrajBERT}, and $\textbf{LLM-SFT}$. The details for these baselines are presented in Appendix~\ref{app:rec-baseline}.

\subsubsection{Evaluation Metrics} 
To evaluate the performances of all compared methods, we adopt two widely used evaluation metrics in existing studies: the top-K Hit Ratio (\textbf{Hit@K}) and the top-K Normalized discounted cumulative gain (\textbf{N@K}). For the next location prediction task, both Hit@K and N@K are utilized, whereas the mobility recovery task is evaluated using 
Hit@K with $K = \{1, 5, 10\}$. Formal definitions of these metrics can be found in Appendix~\ref{app:metric}. We repeat the experiments for 5 times and report the average results for these metrics.

\subsubsection{Parameter Settings}
To derive location tokens, we utilize Llama3.2-1B-Instruct \cite{meta_llama_3_2_1b_instruct} to transform  location descriptions into initial representations. The quantization process is organized into 4 levels, each containing 256 codes of 32 dimensions. 
The model is trained using the AdamW optimizer, with a learning rate of 0.001 and a batch size of 1024. For mobility-specific knowledge infusion, we employ Llama3.2-1B-Instruct as our base model and fine-tune it using Low-Rank Adaptation (LoRA)~\cite{LoRA} with a rank of 128. All location tokens obtained from the quantization stage are appended as new vocabulary entries for the LLM.  
The training process is conducted on four NVIDIA V100 GPUs for 4 epochs. During inference, we apply beam search with a size of 15 to generate multiple candidate predictions. We utilize the \texttt{prefix\_allowed\_tokens\_fn} function from the Transformers library to ensure that the model produces only valid location tokens whenever feasible. For the compared baselines, we adopt the default parameter settings as reported in their respective papers. 

\vspace{-1.5mm}
\subsection{Performance Comparison}

\begin{table*}[thbp]
\centering
\caption{Comparison for next location prediction. The best and second-best results are highlighted in red and green.}
\vspace{-2mm}
\label{tab:next_loc_exp}
\small
\begin{tabularx}{\linewidth}{@{}lXXXXX|XXXXX|XXXXX@{}}
\toprule
Dataset & \multicolumn{5}{c|}{\textbf{NYC}} & \multicolumn{5}{c|}{\textbf{SG}} & \multicolumn{5}{c}{\textbf{CE}} \\ 
\cmidrule(lr){2-16}
 Model & \textbf{Hit@1} & \textbf{Hit@5} & \textbf{Hit@10} & \textbf{N@5} & \textbf{N@10} & \textbf{Hit@1} & \textbf{Hit@5} & \textbf{Hit@10} & \textbf{N@5} & \textbf{N@10} & \textbf{Hit@1} & \textbf{Hit@5} & \textbf{Hit@10} & \textbf{N@5} & \textbf{N@10} \\ 
  \midrule
FPMC           & 0.0852         & 0.2216         & 0.2661          & 0.1577          & 0.1721          & 0.0489         & 0.0966         & 0.1287          & 0.0727          & 0.0830          & 0.0317         & 0.0856         & 0.1219          & 0.0590          & 0.0706          \\
DeepMove       & 0.1108         & 0.2105         & 0.2420          & 0.1635          & 0.1738          & 0.0520         & 0.1131         & 0.1484          & 0.0843          & 0.0956          & 0.0551         & 0.1424         & 0.1979          & 0.1000          & 0.1178          \\
CTLE       & 0.1367         & 0.3031         & 0.3812          & 0.2255          & 0.2473          & 0.0628         & 0.1560         & 0.2147          & 0.1350          & 0.1538          & 0.0728         & 0.1628         & 0.2205          & 0.1232          & 0.1458          \\
TrajFormer       & 0.0693         & 0.1544         & 0.1782          & 0.1283          & 0.1435          & 0.0321         & 0.0638         & 0.0980          & 0.0517          & 0.0634          & 0.0284         & 0.0547         & 0.0942          & 0.0468          & 0.0598          \\
GETNext        & 0.1542         & 0.3347         & 0.4066          & 0.2433          & 0.2668          & 0.0921         & 0.1960         & 0.2509          & 0.1447          & 0.1626          & 0.1029         & 0.2426         & 0.3098          & 0.1763          & 0.1980          \\
MCLP           & 0.1804         & 0.3662         & 0.4356          & 0.2791          & 0.3017          & 0.0839         & 0.1973         & 0.2590          & 0.1422          & 0.1622          & 0.0964         & 0.1929         & 0.2399          & 0.1468          & 0.1620          \\
PLSPL          & 0.1935         & 0.3751         & 0.4507          & 0.2894          & 0.3140          & 0.1055         & 0.2061         & 0.2541          & 0.1571          & 0.1726          & 0.0976         & 0.2261         & 0.2967          & 0.1645          & 0.1872          \\
STHGCN         & 0.1982         & 0.3821         & 0.4469          & \cellcolor{mygreen!25}0.2970          & 0.3180          & \cellcolor{mygreen!25}0.1246         & \cellcolor{mygreen!25}0.2428         & 0.2957          & \cellcolor{mygreen!25}0.1869          & \cellcolor{mygreen!25}0.2041          & 0.0844         & 0.2045         & 0.2669          & 0.1462          & 0.1664          \\
LLM-Move          & 0.1717         & 0.3559         & 0.4397          & 0.2803          & 0.3071          & 0.1010         & 0.1961         & 0.2457          & 0.1513          & 0.1674          & 0.0823         & 0.2078         & 0.2657          & 0.1432          & 0.1629          \\
AgentMove          & 0.1753         & 0.3536         & 0.4377          & 0.2804          & 0.3077          & 0.1118         & 0.2046         & 0.2526          & 0.1558          & 0.1720          & 0.0836         & 0.2093         & 0.2695          & 0.1455          & 0.1650          \\
GenUP          & \cellcolor{mygreen!25}0.2216         & 0.3550         & 0.3784          & 0.2946          & 0.3023          & 0.1227         & 0.2152         & 0.2375          & 0.1726          & 0.1798          & 0.0935         & 0.2008         & 0.2324          & 0.1502          & 0.1605          \\
MobilityLLM    & 0.1840         & \cellcolor{mygreen!25}0.3944         & \cellcolor{myred!25}\textbf{0.4781}          & 0.2949          & \cellcolor{mygreen!25}0.3221          & 0.1077         & 0.2367         & \cellcolor{mygreen!25}0.3045          & 0.1749          & 0.1967          & \cellcolor{mygreen!25}0.1084         & \cellcolor{mygreen!25}0.2623         & \cellcolor{mygreen!25}0.3343          & \cellcolor{mygreen!25}0.1883          & \cellcolor{mygreen!25}0.2116 \\
\midrule
\textbf{QT-Mob}  & \cellcolor{myred!25}\textbf{0.2550} & \cellcolor{myred!25} \textbf{0.4073} & \cellcolor{mygreen!25}0.4562  & \cellcolor{myred!25}\textbf{0.3366}  & \cellcolor{myred!25}\textbf{0.3525}  & \cellcolor{myred!25}\textbf{0.1514} & \cellcolor{myred!25}\textbf{0.2629} & \cellcolor{myred!25}\textbf{0.3171}  & \cellcolor{myred!25}\textbf{0.2105}  & \cellcolor{myred!25}\textbf{0.2281}  & \cellcolor{myred!25}\textbf{0.1243} & \cellcolor{myred!25}\textbf{0.2728} & \cellcolor{myred!25}\textbf{0.3418}  & \cellcolor{myred!25}\textbf{0.2021}  & \cellcolor{myred!25}\textbf{0.2245}  \\
\bottomrule
\end{tabularx}
\end{table*}

\begin{table*}[htbp]
\centering
\small
\caption{\centering Results for mobility recovery task}
\vspace{-2mm}
\begin{tabular}{lccc|ccc|ccc}
\toprule
Dataset & \multicolumn{3}{c|}{\textbf{NYC}} & \multicolumn{3}{c|}{\textbf{SG}} & \multicolumn{3}{c}{\textbf{CE}} \\
\cmidrule(lr){2-10}
     Model           & \textbf{Hit@1} & \textbf{Hit@5} & \textbf{Hit@10} 
                & \textbf{Hit@1} & \textbf{Hit@5} & \textbf{Hit@10} 
                & \textbf{Hit@1} & \textbf{Hit@5} & \textbf{Hit@10} \\
\midrule
LSTM           & 0.1049         & 0.2598         & 0.3115         
               & 0.0583         & 0.1287         & 0.1694    
               & 0.0554         & 0.1463         & 0.2021         \\
AttnMove       & 0.1330         & 0.3061         & 0.3601        
               & 0.0873         & 0.1915         & 0.2456         
               & 0.0582         & 0.1796         & 0.2464         \\
TrajBERT       & 0.1172         & 0.2394         & 0.2654         
               & 0.0664         & 0.1365         & 0.1663         
               & 0.0457   & 0.1280       & 0.1710       \\
LLM-SFT      & 0.1586         & 0.3126         & 0.3625         
               & 0.0897         & 0.1746         & 0.2084         
               & 0.0663         & 0.1744         & 0.2330         \\
\midrule
\textbf{QT-Mob}  & \textbf{0.2284} & \textbf{0.4172} & \textbf{0.4762}
               & \textbf{0.1402} & \textbf{0.2743} & \textbf{0.3331} 
               & \textbf{0.1316} & \textbf{0.2728} & \textbf{0.3321} \\
\bottomrule
\end{tabular}
\label{tab:rec_exp}
\end{table*}

\subsubsection{Next Location Prediction}
The performance of all compared methods across the three datasets is presented in Table \ref{tab:next_loc_exp}. Based on these results, we derive several key observations.

First, among deep learning-based methods, models built on the Transformer architecture demonstrate superior performance. One exception is TrajFormer, which is designed for modeling dense GPS traces with other downstream tasks, failing to tackle next location prediction setting. Besides, FPMC and DeepMove, which rely on statistical models and RNNs, yield low performance across all evaluation metrics. This performance gap highlights the advantages of Transformer-based architectures over RNN structures in modeling sequential data.  However, most Transformer-based models achieve comparable results, with the exception of STHGCN, which exhibits stronger performance, likely due to its ability to capture higher-order dependencies using a hypergraph-based approach.

Second, LLM-based methods exhibit significant potential in understanding mobility data. Even with the zero-shot setting, LLM-Move and AgentMove achieves slighted worse performance than several deep-learning based models. Moreover, GenUP, which converts mobility trajectories into a single templated instruction-tuning dataset for fine-tuning LLMs, outperforms other deep learning models, except STHGCN. Besides, MobilityLLM, which replaces the sequential model component with LLMs, achieves state-of-the-art performance across most metrics among the baselines. These results highlight the superiority of LLMs in capturing mobility insights, both from a textual perspective and as a replacement for traditional deep learning sequential models.

Third, the proposed QT-Mob consistently outperforms all the baselines across most evaluation metrics. For example, QT-Mob achieves substantial improvements in Hit@1, with relative gains of 38.5\%, 40.6\%, and 14.7\% over state-of-the-art MobilityLLM across the three datasets, respectively. Compared to previous LLM-based methods, QT-Mob enhances location representations by replacing conventional location IDs or embeddings with specialized tokens. These tokens not only encode rich semantic information about locations but are also highly compatible with the discrete token structure of LLMs, facilitating more effective integration within the model. Additionally, the incorporation of complementary instruction-tuning datasets further refines the model’s ability to capture complex mobility patterns, contributing to the superior performance of the proposed framework.

\vspace{-2mm}
\subsubsection{Mobility Recovery}

For the mobility recovery task, we have similar observations as in the next location prediction task from Table \ref{tab:rec_exp}. In particular, Transformer-based deep learning models achieves better performance than RNN, and LLM-based method outperforms deep learning models. Notably, our proposed QT-Mob achieves the best performance among all compared methods, indicating its superior ability to comprehend mobility data. This suggests that the enriched semantic representations and fine-tuning strategies incorporated in QT-Mob are not only effective for next location prediction but also generalize well to mobility recovery.

\vspace{-2mm}
\subsection{Model Analysis}

\begin{table}[htbp]
\centering
\small
\caption{\centering Ablation study for
next location prediction.}
\vspace{-2mm}
\begin{tabular}{lccccc}
\toprule
\textbf{NYC} & \textbf{Hit@1} & \textbf{Hit@5} & \textbf{Hit@10} & \textbf{N@5} & \textbf{N@10} \\
\midrule
\textbf{Base}            & 0.2131 & 0.3591 & 0.4084 & 0.2910 & 0.3071  \\
Base+L         & 0.2231 & 0.3624 & 0.4062 & 0.2971 & 0.3114  \\
Base+M         & 0.2249 & 0.3776 & 0.4295 & 0.3057 & 0.3226  \\
Base+U     & 0.2257 & 0.3876 & 0.4347 & 0.3130 & 0.3282  \\
Base+M+U & 0.2361 & 0.4014 & \textbf{0.4603} & 0.3250 & 0.3442  \\
Base+M+L     & 0.2398 & 0.3891 & 0.4329 & 0.3201 & 0.3342  \\
\midrule
\textbf{QT-Mob}    & \textbf{0.2550} & \textbf{0.4073} & 0.4562 & \textbf{0.3366} & \textbf{0.3525} \\
\bottomrule
\end{tabular}
\label{tab:ab_seq_NYC}
\vspace{-3mm}
\end{table}

\begin{table}[ht]
\centering
\small
\caption{\centering Ablation study for mobility recovery.}
\vspace{-2mm}
\resizebox{0.98\linewidth}{!}{
\begin{tabular}{lccc|ccc}
\toprule
\textbf{} & \multicolumn{3}{c|}{\textbf{NYC}} & \multicolumn{3}{c}{\textbf{SG}} \\
\midrule
\textbf{} & \textbf{Hit@1} & \textbf{Hit@5} & \textbf{Hit@10} & \textbf{Hit@1} & \textbf{Hit@5} & \textbf{Hit@10} \\
\midrule
\textbf{Base}    & 0.1586 & 0.3126 &  0.3625 & 0.0897 & 0.1746 & 0.2084  \\
Base+L         & 0.1738 & 0.3295 &  0.3695 & 0.1161 & 0.2267 & 0.2746  \\
Base+N         & 0.2068 & 0.3861 &  0.4420 & 0.1239 & 0.2519 & 0.3041  \\
Base+U     & 0.1947 & 0.3798 &  0.4301 & 0.1260 & 0.2516 & 0.3031  \\
Base+N+U & 0.2195 & 0.4132 & 0.4692 & 0.1376 & 0.2712 & 0.3286  \\
Base+N+L     & 0.2099 & 0.3943 & 0.4472 & 0.1321 & 0.2549 & 0.3132 \\
\midrule
\textbf{QT-Mob}    & \textbf{0.2284} & \textbf{0.4172} & \textbf{0.4762} & \textbf{0.1402} & \textbf{0.2743} & \textbf{0.3331} \\
\bottomrule
\end{tabular}}
\vspace{-3mm}
\label{tab:ab_rec}
\end{table}

\begin{figure}[htbp]
    \centering
    \includegraphics[width=1\linewidth]{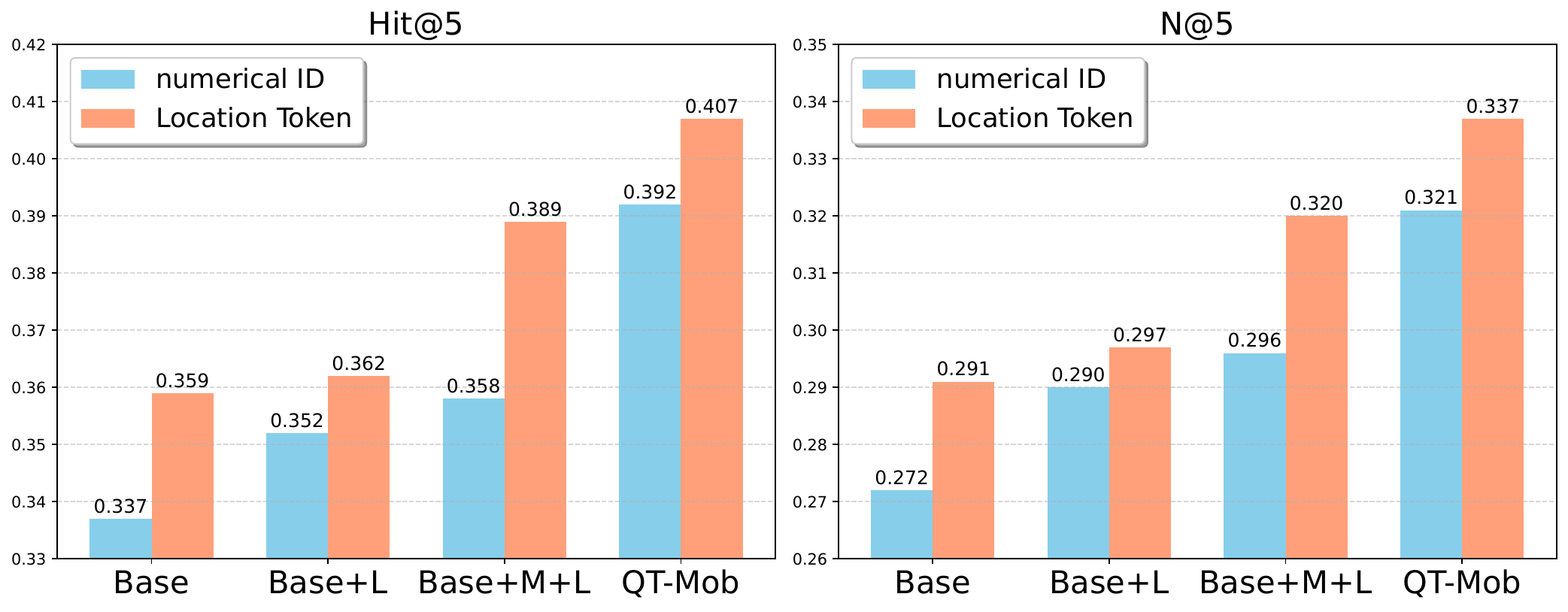}
    \vspace{-7mm}
    \caption{\centering Effect of location tokens on the NYC dataset.}
    \label{fig:POIVsIDNYC}
    \vspace{-3mm}
\end{figure}

\subsubsection{Ablation Study}\label{subsubsec:ablation}
To evaluate the impact of different instruction-tuning components on model performance, we conduct an ablation study by investigating their contributions from the fine-tuning datasets. The datasets consist of four key components: next location prediction (\textbf{+N}) with user profile enhancement (\textbf{+U}), mobility recovery (\textbf{+M}), and location alignment (\textbf{+L}). 

For the \textbf{next location prediction} task, we establish a \textbf{Base} model by performing supervised fine-tuning solely on the next location prediction component (+N). We then examine the performance changes when additional factors (+U, +M, and +L) are incorporated. In contrast, for the \textbf{mobility recovery} task, we establish a \textbf{Base} model using supervised fine-tuning solely on the mobility recovery component (+M) and evaluate the effect of integrating other factors (+U, +N, and +L). The results are presented in Table \ref{tab:ab_seq_NYC} for next location prediction on the NYC dataset and Table~\ref{tab:ab_rec}  for mobility recovery. More experiments on the SG dataset can be found in Appendix~\ref{app:ablation}.

The results indicate that the inclusion of additional instruction-tuning factors consistently improves model performance across all evaluation metrics. This highlights the complementary nature of various components in enhancing LLM comprehension of mobility insights. Notably, the mutual reinforcement of prediction (+N) and recovery (+M) tasks suggests that each task captures distinct yet interrelated dimensions of mobility data, leading to improved downstream performance. Furthermore, location alignment (+L) plays a crucial role by bridging the semantic gap between location tokens and LLM representations, allowing the model to more effectively interpret contextual location semantics and for improved accuracy.  Similarly, adding user profile (+U) provides LLMs with personalized preferences, especially repeated user visits, which is beneficial to the prediction. Furthermore, we observe that when two components are added simultaneously, the model exhibits greater performance gains compared to the addition of a single component. This finding further demonstrates the importance of integrating diverse instruction-tuning components, as adopted in our study.

\subsubsection{Effect of Location Tokens}

We examine the effect of the derived location tokens by replacing them with numerical location IDs (e.g., location 134) as input tokens for LLMs, following the practices adopted in ~\cite{LLMPOI1, GenUP}. Specifically, we evaluate four fine-tuning variants, \textbf{Base}, \textbf{Base+L}, \textbf{Base+M+L}, and \textbf{QT-Mob}, which are defined in Section~\ref{subsubsec:ablation}.

The results, presented in Figure \ref{fig:POIVsIDNYC}, indicate that using the proposed location tokens consistently leads to superior performance across all the cases. This demonstrates that the tokens derived through RQ-VAE effectively capture rich semantics, reflecting both intrinsic attributes and contextual information that discrete numerical location IDs fail to convey. Moreover, all fine-tuning variants benefit from the introduction of semantic location tokens, highlighting the robustness and generalizability of this tokenization approach in mobility analytics.

\subsubsection{Effect of Quantization Process}

\begin{figure}
    \centering
    \includegraphics[width=1\linewidth]{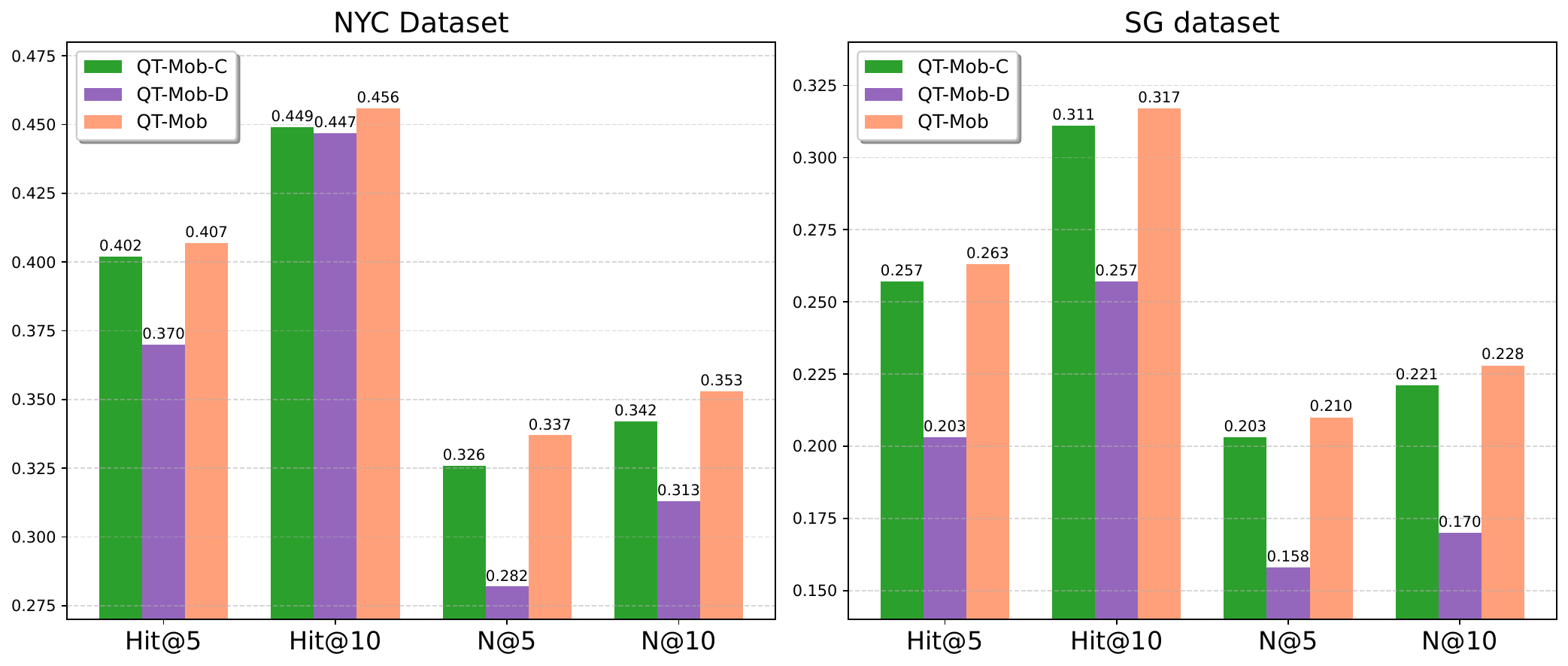}
    \vspace{-7mm}
    \caption{Effect of quantization process.}
    \label{fig:quantization}
    \vspace{-4mm}
\end{figure}

We further evaluate two ablated variants of the model to test the contribution of the proposed quantization process. The first variant, \textbf{QT-Mob-C}, removes the contextual descriptions of each location before quantization, while the second, \textbf{QT-Mob-D},  bypasses the quantization module and directly employs the dense representations produced from the encoder defined in Equation~\ref{eq:encoder} for LLM input after a projection layer. 

The results against QT-Mob on the NYC and the SG dataset are presented in Figure \ref{fig:quantization}. Notably, directly using the latent representations from the encoder results in inferior performance, likely due to a misalignment between the continuous representation space and the discrete processing paradigm of LLMs. Furthermore, the removal of contextual location descriptions leads to performance degradation, indicating that incorporating such contextual information enhances the quality of the quantized location tokens. These findings collectively demonstrate the importance of our quantization strategy.

\begin{table}[htbp]
\centering
\caption{Results for LLM backbones on the NYC dataset}
\vspace{-2mm}
\label{tab:LMvariantNYC}
\small
\begin{tabularx}{\linewidth}{@{}lXXXXX@{}}
\toprule
\textbf{Backbone} & \textbf{Hit@1} & \textbf{Hit@5} & \textbf{Hit@10} & \textbf{N@5} & \textbf{N@10} \\ \midrule
Llama3.2-1B     &  \textbf{0.2550} & 0.4073 & 0.4562  & 0.3366  & 0.3525  \\
Qwen2.5-1.5B       & 0.2524         & 0.4117         & 0.4618          & 0.3394          & 0.3559          \\
TinyLlama-1B           & 0.2449         & 0.4162         & \textbf{0.4688}          & 0.3363          & 0.3535          \\
Phi1.5-1.3B         & 0.2501         & \textbf{0.4169}         & 0.4651          & \textbf{0.3402}          & \textbf{0.3560}          \\ 
OLMo-1B       & 0.2023  & 0.3628  & 0.4073  &  0.2888  & 0.3035  \\
\bottomrule
\end{tabularx}
\end{table}

\begin{figure*}[ht]
    \centering
    \begin{subfigure}{0.24\textwidth}
        \centering
        \includegraphics[width=4.4cm, height=3.4cm]{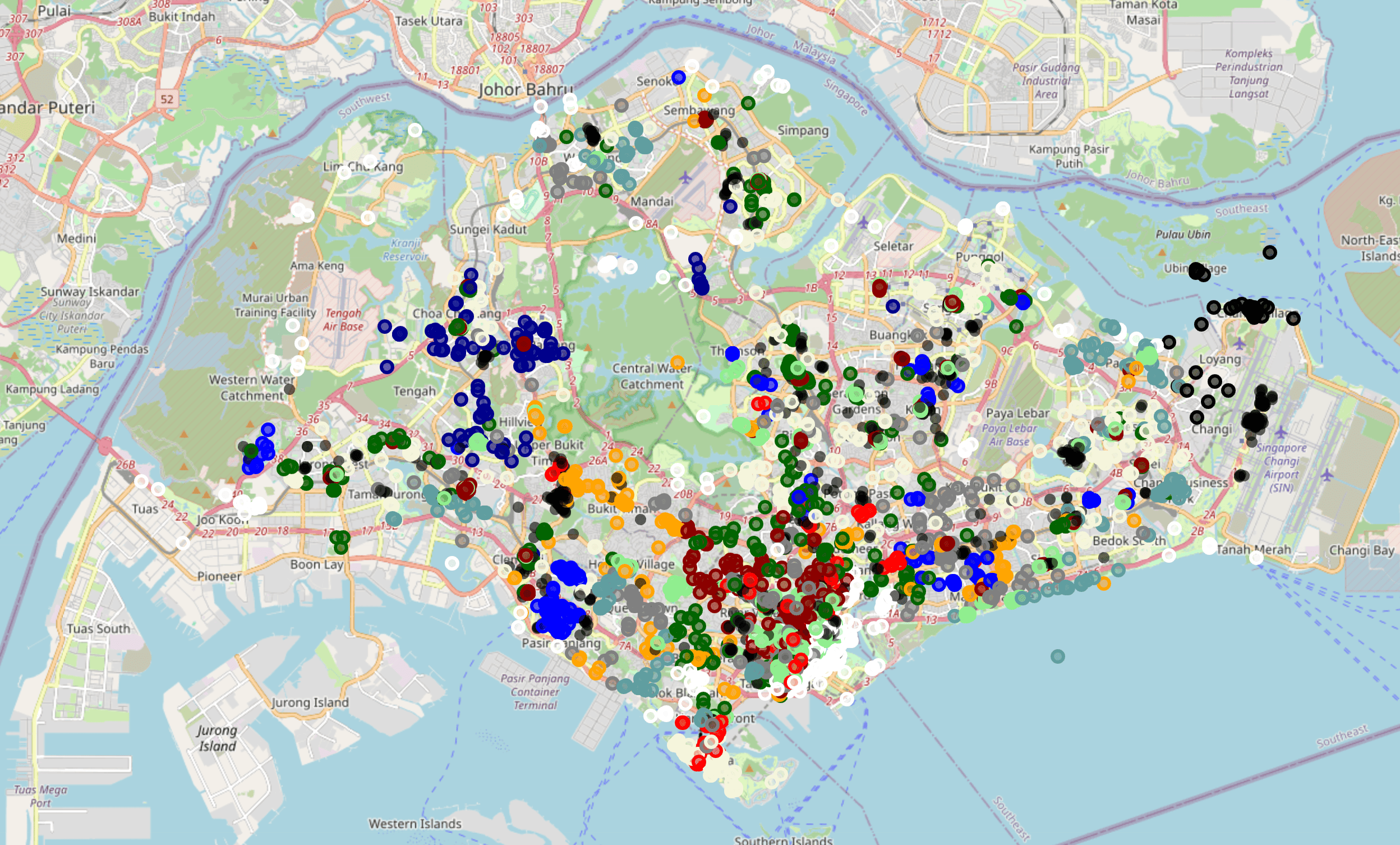}
        \caption{Location tokens in SG}
        \label{subfig:image1}
    \end{subfigure}
    \hfill
    \begin{subfigure}{0.24\textwidth}
        \centering
        \includegraphics[width=4.4cm, height=3.4cm]{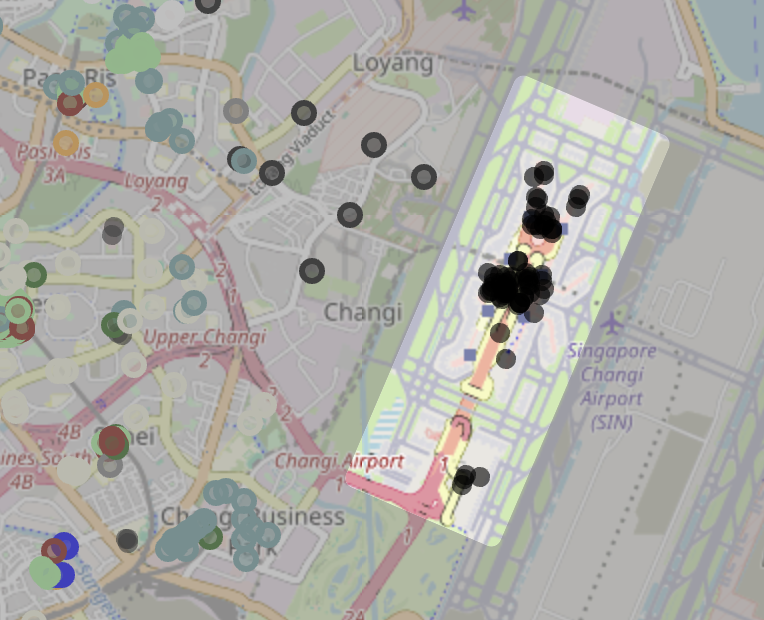}
        \caption{Changi Airport}
        \label{subfig:image2}
    \end{subfigure}
    \hfill
     \begin{subfigure}{0.24\textwidth}
        \centering
        \includegraphics[width=4.4cm, height=3.4cm]{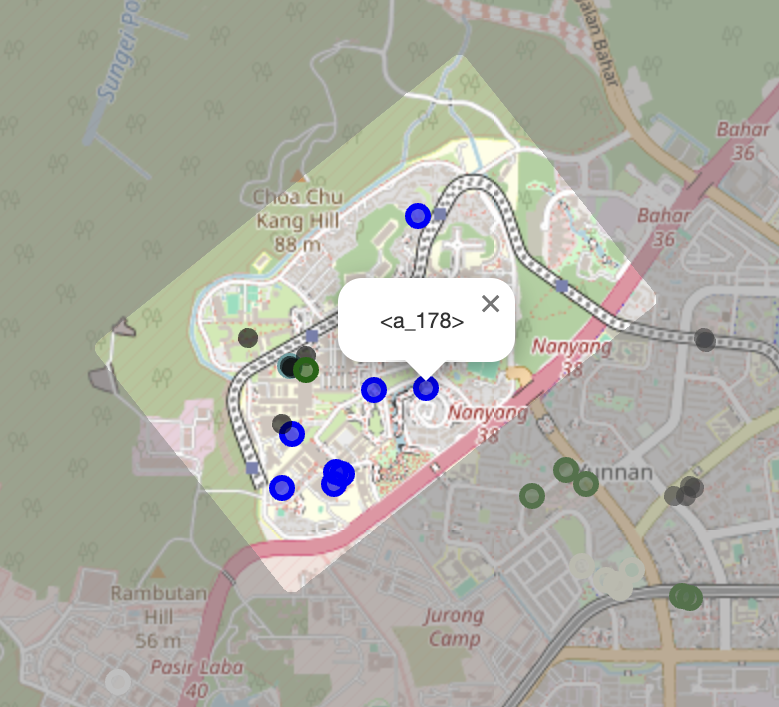}
        \caption{University I}
        \label{subfig:image3}
    \end{subfigure}
    \hfill
    \begin{subfigure}{0.24\textwidth}
        \centering
        \includegraphics[width=4.4cm, height=3.4cm]{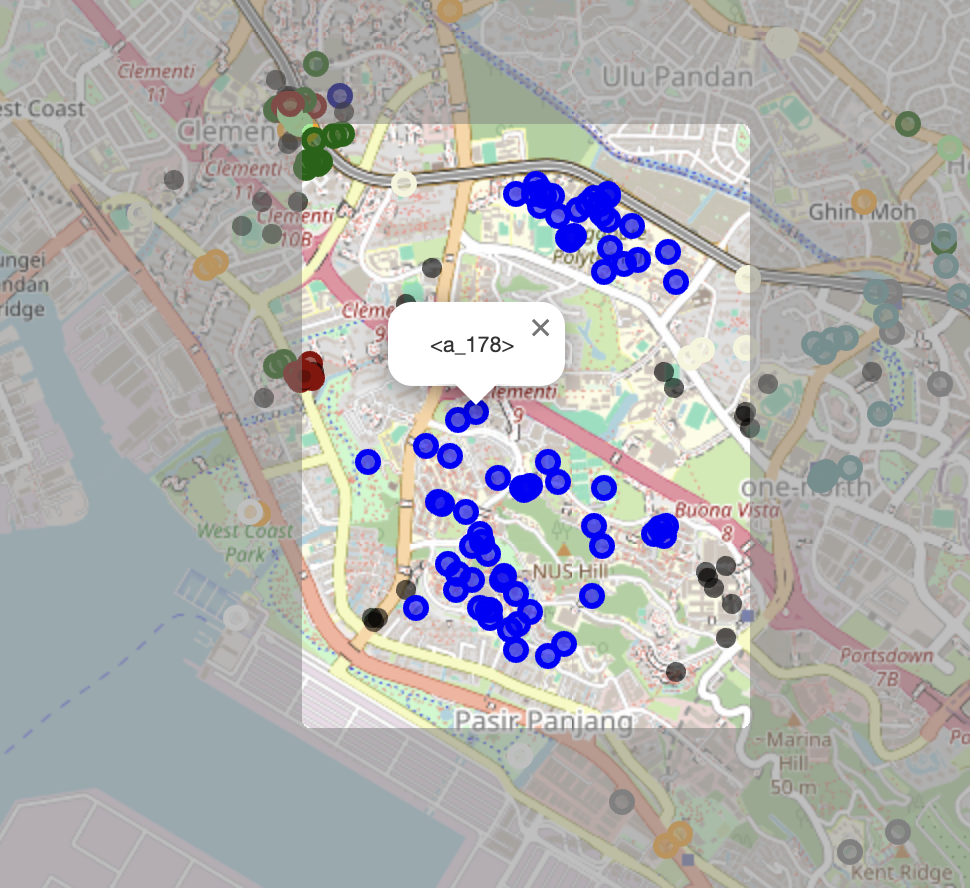}
        \caption{University II}
        \label{subfig:image4}
    \end{subfigure}
    \vspace{-3mm}
    \caption{Visualization of location tokens for the SG dataset}
    \label{fig:visualization}
    \vspace{-4mm}
\end{figure*}

\subsubsection{Effect of Language Models}
To further evaluate the robustness of the proposed QT-Mob framework, we assess the performance using five different LLMs of comparable parameter sizes as the backbone models. In addition to the original Llama3.2-1B, we substitute it with four alternative models: Qwen2.5-1.5B, TinyLlama-1B, Phi1.5-1.3B, and OLMo-1B. The results for the next location prediction task on the NYC and SG datasets are presented in Table~\ref{tab:LMvariantNYC} and Appendix~\ref{app:llm}, respectively.  We observe that, except for OLMo-1B model, the performance of different LLMs remains relatively consistent. This indicates that the QT-Mob framework provides robust and stable performance improvements over existing methods, regardless of the choice of backbone model. Furthermore, no single model demonstrates a clear performance advantage over others, which is likely due to the comparable capabilities of LLMs within similar parameter ranges. Nevertheless, since our framework enhances performance in a model-agnostic manner, the choice of LLM variant does not appear to be a critical factor affecting its effectiveness. This further demonstrates the generalizability and adaptability of QT-Mob across different LLM backbones.

\subsection{Case Study}
To gain an intuition of the semantics encoded in the location tokens, we perform visualizations of these tokens using the SG dataset in Figure~\ref{fig:visualization}.
As the quantization process consists of multiple layers, we focus on the first quantization level, which represents the coarsest granularity, to provide a clear and interpretable analysis.

Figure \ref{subfig:image1} shows a visualization of the entire Singapore region, illustrating the diverse distribution of token representations across different geographic areas. Notably, Figure \ref{subfig:image2} provides a focused view of the Changi Airport area, where all locations within the region share the same token. This token does not appear in locations outside the airport, indicating that it effectively captures both latent semantic properties and geographical characteristics unique to the airport environment. Another example, shown in Figure \ref{subfig:image3} and \ref{subfig:image4}, visualizes locations within two universities in Singapore—one located in a suburban area and the other in downtown. Despite the significant distance between these universities, they share the token $\texttt{<a\_178>}$, suggesting that the tokenization process is influenced not only by spatial proximity but also by semantic similarity across distant locations. 
By leveraging such semantically meaningful and spatially structured location tokens, LLMs can effectively process mobility data using discrete tokenized inputs while maintaining a rich representation of location semantics. 
This enables LLMs to align their internal reasoning capabilities with spatial and contextual factors, ultimately improving performance on downstream mobility analytical tasks.
\section{Conclusion}

In this paper, we introduced QT-Mob, a novel framework designed to enhance the adaptability of LLMs for mobility analytics. Addressing the limitations of existing methods, QT-Mob provides a unified solution for LLMs applied to mobility data, enabling better integration of geospatial knowledge within LLMs. Specifically, QT-Mob integrates a semantic location tokenization module that encodes rich semantic and contextual information of locations into compact location tokens. Furthermore, by incorporating a set of complementary fine-tuning objectives, our approach aligns these representations with the internal structures of LLMs, leading to a more comprehensive understanding of mobility insights. Extensive experiments conducted on three real-world mobility datasets demonstrated the superior performance of QT-Mob in both next location prediction and mobility recovery tasks over existing targeted deep learning and LLM-based methods.

\begin{acks}
The research is supported in part by ST Engineering IHQ Pte. Ltd.; the Ministry of Education, Singapore, under its Academic Research Fund Tier-1 grant RT6/23 and RG101/24; the National Research Foundation, Singapore and DSO National Laboratories under the AI Singapore Programme (AISG Award No. AISG2-RP-2020-019).
\end{acks}

\bibliographystyle{ACM-Reference-Format}
\balance
\bibliography{sample-base}

\appendix
\clearpage
\newpage
\section{Location Description}
A  location description includes its intrinsic attributes, such as name, category, coordinates, and address, as well as  contextual information, such as the top 10 POI categories, top 10 nearest POIs, and top 10 representative nearby POIs, along with their respective distances. A sampled description is shown below:

\label{app:loc-description}
\noindent\fbox{%
    \parbox{0.45\textwidth}{%
    
        The \textbf{name} of this location is \textbf{"Hi-Life Bar \& Grill"} and its \textbf{POI category} is Bar, belonging to the parent category Nightlife Spot.  \\

        The \textbf{geographic coordinates} for this location are (40.785677, -73.976498), with the corresponding geohash code dr72h8gcy9m0.  \\

        The \textbf{address} is 6547 W 83rd St, New York, NY 10024, USA.  \\

\textbf{The top 10 nearby points-of-interest (POI) categories and their counts are:  
}        
        \begin{itemize}
            \item Bar, 47 (avg: 124)  
            \item Coffee Shop, 37 (avg: 63)  
            \item American Restaurant, 28 (avg: 51)  
            \item Office, 27 (avg: 72)  
            \item Café, 23 (avg: 42)  
            \item Metro Station, 19 (avg: 19)  
            \item General Entertainment, 15 (avg: 21)  
            \item Theater, 15 (avg: 19)  
            \item Yoga Studio, 14 (avg: 14)  
            \item Gym, 13 (avg: 25)  
        \end{itemize}

\textbf{The 10 nearest POIs and their distances are:  
}        
        \begin{itemize}
            \item Fred's Restaurant, distance 0.00 km  
            \item Good Enough To Eat, distance 0.01 km  
            \item Hi-Life Bar \& Grill, distance 0.03 km  
            \item Cafe Lalo, distance 0.03 km  
            \item George Keeley's, distance 0.04 km  
            \item Blue Donkey Bar, distance 0.06 km  
            \item Crunch Fitness, distance 0.07 km  
            \item The Dead Poet, distance 0.11 km  
            \item The Gin Mill, distance 0.13 km  
            \item AMC Loews, distance 0.15 km  
        \end{itemize}

\textbf{The 10 representative nearby POIs and their distances are:  
}        
        \begin{itemize}
            \item Metropolitan Museum of Art, distance 1.37 km  
            \item MoMA - Museum of Modern Art, distance 2.70 km  
            \item Apple Store, distance 2.44 km  
            \item Central Park, distance 0.99 km  
            \item Rockefeller Center, distance 2.98 km  
            \item Terminal 5, distance 2.24 km  
            \item American Museum of Natural History, distance 0.60 km  
            \item AMC Loews Lincoln Square 13, distance 1.25 km  
            \item Bloomingdale's, distance 2.69 km  
            \item Radio City Music Hall, distance 2.88 km  
        \end{itemize}
    }%
}

\section{Compared Baselines for Mobility Recovery}
\label{app:rec-baseline}
For mobility recovery task, we compare QT-Mob with the following baselines: 
\begin{itemize}[leftmargin=*]
    \item \textbf{LSTM} \cite{LSTM}: it is a classical RNN variant which processes the incomplete trajectory through an encoder-decoder framework, where the encoder takes in the observed trajectory and the decoder predicts missing locations.
    \item \textbf{AttnMove} \cite{AttnMove}: it is a Transformer-based model that treats missing locations as null tokens in the input and applies inter- and intra-trajectory attention mechanisms to model sequential transition patterns and impute missing locations.
    \item \textbf{TrajBERT} \cite{TrajBERT}: it is a BERT-style model for mobility recovery, employing a bidirectional Transformer encoder to capture bidirectional patterns and incorporating a cross-stage local temporal refinement scheme to enhance accuracy.
    \item \textbf{LLM-SFT}: it fine-tunes LLMs merely for the mobility recovery task, while removing other instruction-tuning tasks described in Section~\ref{subsec:sft}.
\end{itemize}

\section{Evaluation Metrics}
\label{app:metric}

In the experiments, we evaluate the compared methods using  two evaluation metrics, N@$k$ and Hit@$k$.

\paratitle{N@$k$ (Normalized Discounted Cumulative Gain)}:

Normalized Discounted Cumulative Gain (NDCG) is a widely used evaluation metric for ranking systems. Given a list of top-$k$ results, N@$k$ is defined as:  

\[
\text{N}@k = \frac{\text{DCG}@k}{\text{IDCG}@k}
\]

where  

\[
\text{DCG}@k = \sum_{i=1}^{k} \frac{rel_i}{\log_2(i + 1)}
\]

and $rel_i$ is the relevance score of the item at position $i$. The ideal DCG (IDCG) represents the maximum possible DCG value for the list, which is obtained by sorting the list in descending order of relevance.  

\paratitle{Hit@$k$}:

Hit@$k$ is a simple evaluation metric that checks whether there is at least one relevant item in the top-$k$ results. Formally, Hit@$k$ can be expressed as:  

\[
\text{Hit}@k = 
\begin{cases}
1 & \text{if at least one } rel_i > 0 \text{ for } i \in [1, k] \\
0 & \text{otherwise}
\end{cases}
\]

\begin{table}[htb]
\centering
\small
\caption{\centering Ablation study for next location prediction on the SG dataset}
\begin{tabular}{lccccc}
\toprule
\textbf{SG} & \textbf{Hit@1} & \textbf{Hit@5} & \textbf{Hit@10} & \textbf{N@5} & \textbf{N@10} \\
\midrule
\textbf{Base}             & 0.1277 & 0.2257 & 0.2756 & 0.1795 & 0.1956  \\
Base+L         & 0.1366 & 0.2279 & 0.2792 & 0.1846 & 0.2011  \\
Base+M         & 0.1330 & 0.2375 & 0.2890 & 0.1887 & 0.2052  \\
Base+U     & 0.1438 & 0.2603 & 0.3070 & 0.2047 & 0.2199  \\
Base+M+U & 0.1466 & 0.2619 & 0.3149 & 0.2066 & 0.2238  \\
Base+M+L     & 0.1421 & 0.2404 & 0.2864 & 0.1946 & 0.2095  \\
\midrule
\textbf{QT-Mob} & \textbf{0.1514} & \textbf{0.2629} & \textbf{0.3171} & \textbf{0.2105} & \textbf{0.2281} \\
\bottomrule
\end{tabular}
\label{tab:ab_seq_SG}
\end{table}

\section{Additional Experiments}
\subsection{Ablation Study on the SG dataset}\label{app:ablation}

We establish a \textbf{Base} model using supervised fine-tuning solely on the mobility recovery component (+M) and evaluate the effect of integrating other factors (+U, +N, and +L). The results for the next location prediction task on the SG dataset are presented in Table~\ref{tab:ab_seq_SG}. 

Similar to the results on the NYC dataset, the results indicate that different instruction-tuning factors consistently improves model performance across all evaluation metrics. This highlights the complementary nature of various components in enhancing LLM comprehension of mobility insights.

\subsection{Effect of Language Models on the SG dataset}\label{app:llm}

We evaluate the performance using five different LLMs of comparable parameter sizes as the backbone models. In addition to the default Llama3.2-1B, we substitute it with four alternative models: Qwen2.5-1.5B,
TinyLlama-1B,
Phi1.5-1.3B,
and OLMo-1B.
The results for the next location prediction task on the SG datasets are presented in Table~\ref{tab:LMvariantSG}.

\begin{table}[htb]
\centering
\caption{Results for LLM backbones on the SG dataset}
\vspace{-2mm}
\label{tab:LMvariantSG}
\small
\begin{tabularx}{\linewidth}{@{}lXXXXX@{}}
\toprule
\textbf{Backbone} & \textbf{Hit@1} & \textbf{Hit@5} & \textbf{Hit@10} & \textbf{N@5} & \textbf{N@10} \\ \midrule
Llama3.2-1B     & 0.1514 & 0.2629 & 0.3171  & 0.2105  & 0.2281  \\
Qwen2.5-1.5B       & 0.1490         & 0.2643         & 0.3235          & 0.2091          & 0.2282          \\
TinyLlama-1B           & 0.1510  &  \textbf{0.2746}  &  0.3216  &  \textbf{0.2164}  &  0.2316  \\
Phi1.5-1.3B         & \textbf{0.1519}   &  0.2710  &  \textbf{0.3269}   &  0.2143  & \textbf{0.2323} \\ 
OLMo-1B       & 0.1186  & 0.2260  & 0.2749  &  0.1750  &  0.1909 \\
\bottomrule
\end{tabularx}
\end{table}

The results demonstrate that most LLM backbones achieve comparable performance, with slight variations across metrics. The consistently strong performance across different backbones indicates the robustness and generalizability of our proposed framework.

\subsection{Study on Representation Consistency}
To evaluate whether the representations of location descriptions preserve spatial and semantic relationships, we conduct a controlled analysis.  For each location category, we randomly sample one reference location and evaluate its similarity with four groups of comparison samples: 
\begin{itemize}[leftmargin=*]
    \item \textbf{A}/\textbf{B}: 10 nearby/distant locations to evaluate spatial consistency 
    \item \textbf{C}/\textbf{D}: 10  locations from the same category/different categories to evaluate semantic consistency
\end{itemize}

We compute the average cosine similarity between the representation of the reference location and the samples in each group (near vs. distant; same category vs. different category). The results on the NYC and the SG dataset are presented in Table~\ref{tab:consistency}.

\begin{table}[h!]
\centering
\caption{\centering Cosine similarity for different location groups}
\vspace{-2mm}
\begin{tabular}{lcc}
\toprule
       & \textbf{A}/\textbf{B}           & \textbf{C}/\textbf{D}           \\
\midrule
SG     & 0.930 / 0.764 & 0.816 / 0.678 \\
NYC    & 0.937 / 0.780 & 0.832 / 0.680 \\
\bottomrule
\end{tabular}
\label{tab:consistency}
\end{table}

The observed similarity gaps indicate that the quantized representations effectively capture both spatial proximity and semantic relatedness. These results validate the representational consistency across the urban environments.

\subsection{Model Generalization}
Compared to previous baselines, our model indeed has the potential to tackle unseen locations. To evaluate this, we randomly sample 100 locations in test data and remove the trajectories that contain them in the training dataset. As most baselines cannot handle such cases, we compare it with LLM-Move~\cite{LLMprompt2} following its zero-shot ranking setting. The results, shown in Table~\ref{tab:unseen}, highlight the generalization capability of QT-Mob in these scenarios.

\begin{table}[htb]
\centering
\caption{Results for model generalization}
\vspace{-2mm}
\label{tab:unseen}
\small
\begin{tabularx}{\linewidth}{lXXXXX}
\toprule
\textbf{} & \textbf{Hit@1} & \textbf{Hit@5} & \textbf{Hit@10} & \textbf{N@5} & \textbf{N@10} \\ \midrule
LLM-Move     & 0.290 & 0.340 & 0.420  & 0.327  & 0.344  \\
QT-Mob       & \textbf{0.320}         & \textbf{0.380}         & \textbf{0.480}          & \textbf{0.348}          & \textbf{0.379} \\ 
\bottomrule
\end{tabularx}
\end{table}

\subsection{Parameter Study}
To study the sensitivity of the model parameters, we conduct a parameter study by varying the number of quantization levels ($L$) and the codebook size ($K$) at each layer. Due to space limit, we report the results on the NYC dataset for next location prediction in Table~\ref{tab:k_variants} and Table~\ref{tab:l_variants}.  Similar trends can be observed in other datasets.

\begin{table}[h!]
\centering
\caption{\centering Performance under different $K$ values $(L=4)$}
\vspace{-2mm}
\begin{tabularx}{\linewidth}{lXXXXX}
\toprule
\textbf{$K$} & \textbf{Hit@1} & \textbf{Hit@5} & \textbf{Hit@10} & \textbf{N@5} & \textbf{N@10} \\
\midrule
64  & 0.2394 & 0.4110 & 0.4558 & 0.3314 & 0.3461 \\
128 & 0.2401 & 0.4036 & 0.4555 & 0.3283 & 0.3454 \\
256 (QT-Mob) & 0.2550 & 0.4073 & 0.4562 & 0.3366 & 0.3525 \\
512 & 0.2457 & 0.4099 & 0.4570 & 0.3336 & 0.3490 \\
\bottomrule
\end{tabularx}
\label{tab:k_variants}
\end{table}

\begin{table}[h!]
\centering
\caption{\centering Performance under different $L$ values $(K=256)$}
\vspace{-2mm}
\begin{tabularx}{\linewidth}{lXXXXX}
\toprule
\textbf{$L$} & \textbf{Hit@1} & \textbf{Hit@5} & \textbf{Hit@10} & \textbf{N@5} & \textbf{N@10} \\
\midrule
2 & 0.2301 & 0.4184 & 0.4822 & 0.3312 & 0.3518 \\
3 & 0.2390 & 0.4084 & 0.4581 & 0.3295 & 0.3456 \\
4 (QT-Mob) & 0.2550 & 0.4073 & 0.4562 & 0.3366 & 0.3525 \\
5 & 0.2342 & 0.4125 & 0.4610 & 0.3301 & 0.3460 \\
\bottomrule
\end{tabularx}
\label{tab:l_variants}
\end{table}

The results indicate that the default configuration used in QT-Mob ($K=256$, $L=4$)  achieves strong performance across all metrics. Additionally, the performance variations across different parameter settings are relatively minor. This shows the stability of the proposed model architecture.

\end{document}